\documentclass[3p,11pt]{elsarticle}

\usepackage{amsmath}
\usepackage{amssymb}
\usepackage{bm}
\usepackage{booktabs}
\usepackage{graphicx}
\usepackage{multirow}
\usepackage{array}
\usepackage{xcolor}
\usepackage[ruled,vlined,linesnumbered]{algorithm2e}
\usepackage{url}
\usepackage[hidelinks]{hyperref}

\newcommand{\notavail}{\textemdash}

\newcommand{\best}[1]{\textbf{#1}}

\journal{arXiv preprint}

\begin{document}

\begin{frontmatter}

\title{Revisiting Multi-Object Tracking Baselines:
       Hyperparameter Optimization with Multi-Fidelity Greedy Coordinate Search}

\author[1]{Momir Ad\v{z}emovi\'{c}}
\ead{pd222011@alas.matf.bg.ac.rs}

\affiliation[1]{organization={Department of Computer Science, Faculty of Mathematics, University of Belgrade},
                addressline={Studentski trg 16},
                city={Belgrade},
                postcode={11000},
                country={Serbia}}

\begin{abstract}
Multi-object tracking (MOT) is dominated by the tracking-by-detection paradigm, whose
methods typically rely on a small set of hyperparameters that are conventionally chosen by
hand. Tuning them requires repeated expert-guided experimentation, while the procedures
used to select reported values are often not systematically evaluated or fully documented.
Hyperparameter optimization (HPO) automates this process, yet it remains rarely used in
MOT, and existing studies applying HPO to MOT predate modern deep-detector-based trackers
and HOTA evaluation. We systematically apply HPO across two datasets and four
tracking-by-detection methods. We also propose Multi-Fidelity Greedy Coordinate Search
(MFGCS), which optimizes one hyperparameter at a time by first evaluating candidate values
on a small subset of scenes and re-evaluating only promising candidates on the full
dataset. Across all eight tracker–dataset combinations, the Tree-structured Parzen
Estimator (TPE) and MFGCS outperform both our hand-tuned configurations and the
corresponding published results, with improvements of up to $4.38$ and $16.05$ HOTA
points, respectively. MFGCS also reaches a predefined HOTA target faster than TPE in seven
of the eight combinations. Within each tracker–dataset pair, all optimizers share the same
search space and evaluation pipeline, isolating the effect of the search strategy. We
release the code and tuned configurations to enable future work to compare against
systematically optimized rather than default or manually tuned baselines.
\end{abstract}


\end{frontmatter}

\section{Introduction}
\label{sec:introduction}

Multi-object tracking (MOT) aims to localize objects in a video while maintaining their
identities over time. It is a core component of video-based perception in autonomous
driving \citep{geiger2012kitti}, robotics \citep{xu2024robotnav}, sports analytics
\citep{cui2023sportsmot}, surveillance \citep{urbann2021surveillance}, animal behaviour
analysis \citep{cao2025topic}, and retail analytics \citep{adzemovic2025survey}. The
dominant paradigm is tracking-by-detection, in which an object detector is applied
independently to each frame and a separate association step links the resulting detections
to existing tracks \citep{bewley2016sort, zhang2022bytetrack}. Its main advantage is
modularity: the detector and association pipeline can be developed and replaced
independently, allowing a stronger detector to be substituted without retraining the
tracker. However, the association stage remains largely based on hand-designed methods
whose complexity has increased over time.

The pioneering SORT method combines a Kalman filter with Intersection-over-Union matching
\citep{bewley2016sort}, while Deep SORT adds appearance embeddings to the association cost
\citep{wojke2017deepsort}. Later methods incorporate further mechanisms, including a
second matching pass over low-confidence detections \citep{zhang2022bytetrack},
camera-motion compensation, observation-centric motion modelling, pseudo-depth ordering,
and detection-confidence modulation \citep{aharon2022botsort, cao2023ocsort,
liu2023sparsetrack, yang2024hybridsort, adzemovic2025deepmovesort}. While these mechanisms
can improve tracking performance, they also introduce additional hyperparameters that must
be manually specified or tuned. Common examples include detection and association
thresholds, gating margins, and track lifecycle parameters, with each additional
association cue introducing further settings. As a result, even relatively simple
tracking-by-detection methods can depend substantially on how their hyperparameters are
selected.

In practice, selecting these hyperparameters often requires substantial manual tuning
\citep{stanczyk2025mcbyte}. For the trackers evaluated in this study, manual tuning
required several hours to approximately one day per tracker. Moreover, the benchmark
results reported for the considered methods \citep{bewley2016sort, zhang2022bytetrack,
liu2023sparsetrack, adzemovic2024movesort} generally provide little detail about the
procedure used to select their hyperparameters, making the amount of tuning effort
difficult to assess. This creates two problems. First, comparisons between a new tracker
and its predecessor may conflate the contribution of the proposed component with
differences in tuning effort. Second, a configuration selected on one benchmark may
transfer poorly to another. This issue is particularly relevant for DanceTrack
\citep{sun2022dancetrack} and SportsMOT \citep{cui2023sportsmot}, whose motion and
appearance characteristics differ substantially from those of the benchmarks on which many
tracking-by-detection methods were originally developed. We therefore use these datasets
to study the effect of systematic retuning and find substantial performance gains from
optimizing tracker hyperparameters separately for each benchmark.

These issues motivate treating tracker tuning as a hyperparameter optimization (HPO)
problem, replacing the manual tuning loop with an automated search within a predefined
search space and evaluation budget \citep{bergstra2012random, feurer2019hpo}. MOT is
particularly well suited to this formulation because tracker performance is effectively a
black-box objective with respect to its hyperparameters, as the discrete tracking and
evaluation pipeline provides no usable gradient for optimization. Evaluations are also
expensive, while performance can be estimated using different amounts of data, making the
evaluated data subset a natural source of fidelity. Despite this fit, HPO remains rarely
used in MOT. When applied, prior work typically considers a single search strategy on a
single tracking domain or benchmark \citep{fleck2021tuningmot, lurbur2026bycatch}, and the
one study that does compare several optimizers \citep{madrigal2019hpotools} predates
deep-detector-based trackers and HOTA-based evaluation, as detailed in
Section~\ref{sec:related_work}.

In this work, we systematically study the choice of optimization strategy for MOT
hyperparameter tuning. We first compare eight optimization methods spanning six families
under a common search space using SORT on DanceTrack. The two strongest methods are then
evaluated across two datasets, DanceTrack and SportsMOT, and four tracking-by-detection
methods: SORT \citep{bewley2016sort}, ByteTrack \citep{zhang2022bytetrack}, SparseTrack
\citep{liu2023sparsetrack}, and MoveSORT-KF \citep{adzemovic2024movesort}. Each tracker
has its own search space, which is shared by all optimization methods used to tune it. We
also propose Multi-Fidelity Greedy Coordinate Search (MFGCS), an optimization method
designed to reduce the cost of full-dataset evaluations. MFGCS tunes one hyperparameter at
a time, first comparing candidate values on a small subset of scenes to obtain a cheap but
noisy estimate of their relative performance. The full dataset is evaluated only for a
proposed update, which is accepted only if it improves the full-dataset score.

Our contributions are as follows:

\begin{itemize}

  \item We introduce Multi-Fidelity Greedy Coordinate Search (MFGCS), a coordinate-wise
        optimizer that first compares candidate values using inexpensive evaluations on
        scene subsets and evaluates the complete dataset only for promising updates. A
        change is accepted only if it improves the complete-dataset objective. We further
        ablate its three interchangeable components.
      
  \item We provide a systematic study of hyperparameter optimization for modern
        tracking-by-detection MOT. We compare eight optimizers from six families under a
        controlled search space, and evaluate the strongest methods, MFGCS and TPE, across
        two datasets and four trackers. To our knowledge, this is the first broad
        comparison of HPO strategies for modern tracking-by-detection methods, the first
        using HOTA as the optimization objective, and the first to evaluate multi-fidelity
        HPO in this setting.
      
  \item We show that commonly used tracker configurations substantially underestimate the
        performance of existing baselines. MFGCS and TPE improve the manually tuned
        configuration in all eight tracker--dataset combinations, with gains of up to
        $4.38$ HOTA points, and retuning reduces the apparent improvement of newer
        trackers over their predecessors. These results show that conclusions about
        architectural progress can be confounded by differences in hyperparameter tuning.
      
  \item We verify that these gains transfer to held-out test data: for every
        tracker--dataset pair, the better validation-selected configuration exceeds the
        corresponding published test result, with improvements of up to $8.69$ HOTA
        points. We release the code and optimized configurations so that future work can
        compare against systematically tuned rather than default or manually selected
        baselines.
      
\end{itemize}
      
\section{Related work}
\label{sec:related_work}

We review three lines of related work. First, we cover black-box hyperparameter
optimization, which encompasses the main optimizer families evaluated in this study.
Second, we discuss multi-fidelity hyperparameter optimization, the line of work most
closely related to MFGCS. Finally, we review prior applications of hyperparameter
optimization to multi-object tracking.

\textbf{Black-box hyperparameter optimization.} Non-model-based methods explore the search
space without constructing an explicit model of the objective. Random search is a standard
baseline and was empirically shown to match or outperform grid search on the search spaces
studied by \citet{bergstra2012random}. Model-based methods instead fit a surrogate to
previous evaluations and use it to select promising candidates
\citep{snoek2012practical,hutter2011smac,feurer2019hpo}. The Tree-structured Parzen
Estimator (TPE) models the densities of well-performing and remaining configurations
separately and favors candidates that are more likely under the former
\citep{bergstra2011tpe}. Gaussian-process Bayesian optimization (GP-BO) places a posterior
over the objective and selects candidates by optimizing an acquisition function
\citep{snoek2012practical}, while SMAC replaces the Gaussian process with a random forest,
making it well suited to mixed and conditional search spaces \citep{hutter2011smac}. In
our Bayesian optimization experiments, we use expected improvement, which scores
candidates according to their expected improvement over the best observed value. A
separate class of derivative-free methods is direct search, including coordinate descent
\citep{wright2015coordinate} and pattern search \citep{kolda2003directsearch}, which
optimize the objective without gradients or a learned surrogate by systematically
exploring local changes in the search space. We represent this family with Greedy
Coordinate Search (GCS), the single-fidelity coordinate-wise baseline for MFGCS. The other
black-box methods evaluated in this work are random search, TPE, and GP-BO.

\textbf{Multi-fidelity optimization.} When a cheaper approximation of the objective is
available, the evaluation budget can be allocated adaptively across fidelity levels. Many
configurations can first be evaluated at low fidelity, with additional resources assigned
only to the more promising ones. Successive Halving follows this principle by evaluating a
large set of configurations with a small resource allocation and repeatedly promoting a
fraction of the best-performing configurations to higher fidelities
\citep{jamieson2016successive}. Hyperband runs multiple Successive Halving brackets with
different trade-offs between the number of configurations and the resources allocated to
each \citep{li2018hyperband}. BOHB replaces Hyperband's random configuration sampling with
a TPE-style model \citep{falkner2018bohb}, while ASHA adapts Successive Halving to
asynchronous parallel execution \citep{li2020asha}. Unlike model training, tracker
hyperparameter evaluation does not naturally provide epochs or training iterations as a
fidelity variable. We therefore define fidelity by the amount of data evaluated, using
subsets of scenes for lower-fidelity evaluations and the complete dataset for the highest
fidelity. We evaluate Hyperband and BOHB using this fidelity axis. MFGCS uses the same
notion of fidelity but applies it within coordinate-wise search: low-fidelity evaluations
compare candidate values for a single hyperparameter around the current configuration,
while the complete dataset determines whether the proposed update is accepted. In our
experiments, MFGCS outperforms both standard Hyperband and BOHB under this common fidelity
definition.

\textbf{Hyperparameter optimization applied to MOT.} Applications of hyperparameter
optimization to MOT remain sparse. To our knowledge, the only prior study that directly
compares multiple optimizers is \citet{madrigal2019hpotools}, which evaluates MCMC, SMAC,
TPE, and Spearmint on PETS09 and ETH. That work predates the current generation of
deep-detector-based trackers and HOTA-based evaluation, and does not consider
multi-fidelity or direct-search methods. More recent studies have used HOTA directly as a
tuning objective on a UA-DETRAC sequence \citep{mohamed2022pae}, in the winning entry of
the Fish Tracking Challenge 2024 \citep{itoh2024fishchallenge}, and on a trawl-camera
dataset \citep{lurbur2026bycatch}, but each applies a single search strategy within a
single domain. The closest prior study to ours is \citet{lurbur2026bycatch}, which also
tunes four trackers using HOTA, but does so with one optimizer on one domain-specific
dataset. In contrast, we systematically compare multiple HPO strategies, including
multi-fidelity and direct-search methods, across several modern tracking-by-detection
trackers and two public datasets under a common evaluation protocol.

\section{Tracker hyperparameter optimization}
\label{sec:tracker_hpo}
\label{sec:problem_formulation}
\label{sec:budget}

We formulate tracker tuning as a single-objective black-box optimization problem and
define a common budget unit for comparing optimizers throughout the paper. A common budget
is necessary because the evaluated methods can incur substantially different costs per
evaluation, making the number of evaluations alone an unsuitable basis for comparison.

Let $D = \{s_{1}, s_{2}, \dots, s_{|D|}\}$ denote a dataset split consisting of $|D|$
scenes, and let $\bm{\theta} \in \Theta$ denote a vector of tracker hyperparameters from a
bounded search space $\Theta$. We define $f(\bm{\theta}; D)$ as the tracking performance
obtained by evaluating the tracker configured with $\bm{\theta}$ on $D$; the metric used
throughout this work is defined in Section~\ref{sec:setup}. The optimization problem is

\begin{equation}
\bm{\theta}^{*} = \arg\max_{\bm{\theta} \in \Theta} f(\bm{\theta}; D),
\label{eq:objective}
\end{equation}

where $\bm{\theta}^{*}$ denotes the optimal configuration in the search space. The
objective is treated as a black box because the discrete tracking and evaluation pipeline
provides no usable gradient with respect to the tracker hyperparameters.

Optimizers that evaluate candidates on subsets of scenes incur different costs per trial,
making trial counts unsuitable for comparing computational effort. We therefore use the
\emph{cumulative number of scenes evaluated} as a common cost measure throughout the
paper,
\begin{equation}
B = \sum_{i=1}^{N} |D_i|,
\label{eq:budget}
\end{equation}
where $i$ indexes every objective evaluation performed during an optimization run and $D_i
\subseteq D$ is the set of scenes used in that evaluation. Thus, every evaluation
contributes to $B$, including the initial evaluation, low-fidelity candidate evaluations,
and full-dataset evaluations, regardless of whether it results in an accepted update or a
completed optimizer trial. Because a common trial count does not imply equal computational
cost across optimizers, we report the cumulative number of scenes actually evaluated by
each run. This measure assumes that scenes within a dataset have approximately similar
lengths. When scene lengths differ substantially, the cumulative number of evaluated
frames would provide a more accurate measure of computational effort.

\section{Multi-Fidelity Greedy Coordinate Search}
\label{sec:mfgcs}

We designed MFGCS around two characteristics of the tracker-tuning problem described in
Section~\ref{sec:tracker_hpo}. First, single-fidelity methods such as TPE, random search,
and GP-BO evaluate every proposed candidate on the complete dataset split, including
candidates that are ultimately unpromising. MFGCS instead compares candidate values on a
small subset of scenes and performs a complete-dataset evaluation only for candidates that
pass this initial comparison. This is intended to reach comparable tracking performance
while requiring substantially fewer evaluated scenes.

The second property concerns the reliability of low-fidelity evaluations. A score measured
on a small subset of scenes can be a noisy estimate of performance on the complete split,
with much of the variation arising from which scenes are sampled. MFGCS therefore
evaluates all candidate values for a given hyperparameter on the same scene subset. This
controls for scene-selection effects and makes their relative performance more
informative, even when the absolute low-fidelity scores are inaccurate. The subset
evaluation is therefore used only to identify the most promising candidate; an update is
accepted only after it improves the score on the complete dataset.

We further assume that the objective exhibits useful local structure along individual
coordinates, such that varying one hyperparameter while holding the others fixed can
reveal consistent improvements. This assumption is only approximate because tracker
hyperparameters interact, but in practice the objective is sufficiently structured for
coordinate-wise search to provide an effective optimization strategy.

\subsection{Algorithm}
\label{sec:mfgcs_algorithm}

Algorithm~\ref{alg:mfgcs} summarizes the procedure. The outer loop sweeps over the
coordinates of the hyperparameter vector $\bm{\theta} \in \Theta$. For each coordinate,
the scene sampler selects a subset $D_s$ of $m$ scenes, and a coordinate optimizer
searches the selected hyperparameter on $D_s$ while holding all remaining coordinates
fixed. The search is anchored at the current value, so if no candidate improves upon it on
the subset, the coordinate remains unchanged and the full-dataset evaluation is skipped.
Otherwise, the proposed candidate is accepted only if it improves the full-dataset score
by more than $\varepsilon$. Thus, the subset evaluation determines which candidates are
considered for an update, while the full-dataset evaluation determines whether the update
is accepted.

\begin{algorithm}[t]
\SetAlgoLined
\DontPrintSemicolon
\KwIn{search space $\Theta$, dataset $D$, scene sampler $S$, coordinate optimizer $C$,
      max sweeps $T$, subset size $m$, acceptance threshold $\varepsilon$}
\KwOut{configuration $\bm{\theta}$ and its full-dataset score $v$}
$\bm{\theta} \leftarrow$ initial configuration\;
$v \leftarrow f(\bm{\theta}; D)$ \tcp*{bootstrap full-dataset score}
\For{$t \leftarrow 1$ \KwTo $T$}{
  $\mathit{improved} \leftarrow \textsc{False}$\;
  \ForEach{coordinate $p$ of $\bm{\theta}$}{
    $D_s \leftarrow S(D, m)$ \tcp*{low-fidelity scene subset}
    $\bm{\theta}' \leftarrow C(\bm{\theta}, p, D_s)$ \tcp*{search one coordinate}
    \If{$\theta'_p = \theta_p$}{
      \textbf{continue} \tcp*{anchored; skip the doomed full evaluation}
    }
    $v' \leftarrow f(\bm{\theta}'; D)$ \tcp*{high-fidelity acceptance gate}
    \If{$v' > v + \varepsilon$}{
      $\bm{\theta}, v \leftarrow \bm{\theta}', v'$\;
      shrink $\Theta_p$ around $\theta_p$\;
      $\mathit{improved} \leftarrow \textsc{True}$\;
    }
  }
  \lIf{$\lnot\,\mathit{improved}$}{\textbf{break}}
}
\Return $\bm{\theta}, v$\;
\caption{Multi-Fidelity Greedy Coordinate Search}
\label{alg:mfgcs}
\end{algorithm}

\subsection{Components}
\label{sec:mfgcs_components}

Algorithm~\ref{alg:mfgcs} comprises three configurable components: a scene sampler, a
coordinate optimizer, and a per-coordinate early-stopping rule. We describe our choices below.
Section~\ref{sec:ablations} ablates the scene sampler and the coordinate optimizer; the
early-stopping rule is held fixed at the setting given here throughout our experiments.

\textbf{Scene sampler.} The scene sampler selects a subset $D_s \subset D$ by sampling $m$
scenes uniformly without replacement. All candidate values for a given coordinate are
evaluated on the same subset, and only the proposed value returned by the coordinate
optimizer is evaluated on the full dataset. Using a common subset is important because it
keeps the low-fidelity comparison between candidate values consistent.

\begin{figure}[t]
\centering
\includegraphics[width=0.82\textwidth]{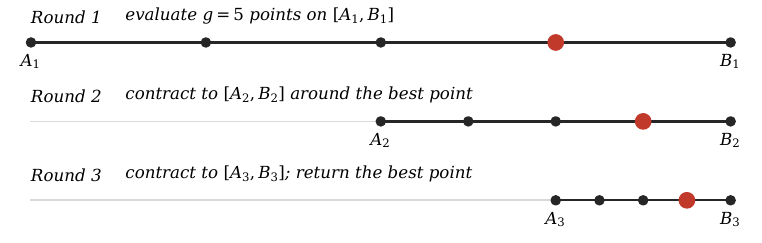}
\caption{Coarse-to-fine grid search on one continuous coordinate ($g = 5$, $r = 3$). Each
round keeps the best point (red) and contracts around it, giving $[A_{1}, B_{1}] \supset
[A_{2}, B_{2}] \supset [A_{3}, B_{3}]$. The best point of the final round supplies the
coordinate's value in the candidate configuration that is then evaluated on the full
dataset.}
\label{fig:grid_contraction}
\end{figure}

\textbf{Coordinate optimizer.} The coordinate optimizer searches a single hyperparameter
and returns a candidate for full-dataset evaluation. Continuous parameters are optimized
using coarse-to-fine grid search: the optimizer evaluates $g$ evenly spaced points on the
current interval $[A,B]$, including both endpoints, retains the best-performing value,
contracts the interval around it, and repeats the procedure for $r$ rounds.
Figure~\ref{fig:grid_contraction} illustrates one such search. Integer-valued parameters
use the same procedure, with each candidate rounded to the lattice and duplicate values
evaluated only once. Categorical parameters have no natural ordering suitable for interval
refinement, so the optimizer enumerates all levels instead. After a numeric value
$\hat{\theta}_{p}$ is accepted on an interval $[A,B]$ of width $w$, we narrow the interval
to $[\hat{\theta}_{p}-\rho w,\,\hat{\theta}_{p}+\rho w]$, clipped to $[A,B]$, with integer
endpoints re-rounded to the lattice when necessary. Categorical search spaces are not
narrowed. We compare this grid search against ternary-section and random-candidate
alternatives in Section~\ref{sec:ablations}, which describes all three, and
Appendix~\ref{app:ternary_vs_grid} derives the trade-off between grid and ternary
formally.

\textbf{Per-coordinate early stopping.} A coordinate is removed from subsequent sweeps
after producing no accepted update for two consecutive sweeps, concentrating the remaining
evaluations on coordinates that continue to yield improvements. The outer loop terminates
when an entire sweep produces no accepted update, when the sweep limit $T$ is reached, or
when the full-dataset evaluation budget is exhausted. The bootstrap evaluation counts
toward this budget.

\section{Experimental evaluation}
\label{sec:experiments}

We evaluate hyperparameter optimization on two multi-object tracking datasets, DanceTrack
\citep{sun2022dancetrack} and SportsMOT \citep{cui2023sportsmot}, using four
tracking-by-detection methods. We first describe the experimental setup, then identify the
two optimization methods used in the full study, present the main results, and conclude
with ablation experiments.

\subsection{Experimental setup}
\label{sec:setup}
\label{sec:benchmarks}

\textbf{Hardware and software.} All runs used a single machine with a 12th-generation
Intel
Core i7-12700K CPU and an NVIDIA GeForce RTX 3070 GPU. All experiments were run with the
Motrack library, which is publicly available at
\url{https://github.com/Robotmurlock/Motrack} together with the optimized configurations
of Tables~\ref{tab:best_sort_movesort} and
\ref{tab:best_byte_sparse}. Every run uses seed $42$.

\textbf{Datasets.} DanceTrack\footnote{DanceTrack GitHub page:
\url{https://github.com/DanceTrack/DanceTrack}.} consists of group dance recordings in
which
the targets share a nearly identical appearance and move non-linearly, with frequent
crossings and re-entries \citep{sun2022dancetrack}. Appearance cues are therefore largely
uninformative, so the benchmark isolates the association component of a tracker. Its
validation split has $25$ scenes. SportsMOT\footnote{SportsMOT GitHub page:
\url{https://github.com/MCG-NJU/SportsMOT}.} consists of basketball, football, and
volleyball
sequences recorded with a moving camera \citep{cui2023sportsmot}, where targets move fast
and
irregularly and camera motion adds apparent displacement that the motion model must
absorb.
Its validation split has $45$ scenes and its test split is more than three times that
size.
We run every search on the validation splits, and use the test splits only for the final
evaluation.

\textbf{Evaluation metrics.} We follow the TrackEval definitions \citep{luiten2021hota}
and report HOTA together with its detection accuracy (DetA) and association accuracy
(AssA) components, as well as MOTA \citep{bernardin2008clear}, IDF1
\citep{ristani2016performance}, and the number of identity switches (IDSW). We use HOTA as
the optimization objective because it jointly captures detection and association
performance through a geometric-mean formulation, providing a single scalar objective
without introducing an explicit weighting between the two. We additionally report
association throughput for each tracker in Appendix~\ref{app:full_metrics}, but do not
include it in the optimization objective. With cached detections and the hardware
described above, the association stage runs at $698$--$1121$ frames per second.

\textbf{Optimizer metrics.} We compare optimization runs using the cumulative scene budget
defined in Equation~\ref{eq:budget}, which accounts for all evaluations performed and
provides a hardware-independent measure of search effort. We report wall-clock time only
as \emph{time-to-target}, defined as the time at which the running-best HOTA first reaches
a predefined target. This metric is reported only in Table~\ref{tab:main_results}, where
all eight runs use the same execution configuration and are therefore directly comparable.
Time-to-target measures the cost of reaching a useful configuration rather than completing
the entire search, which is important because an optimizer that finds a strong
configuration early can be terminated early.

\textbf{Trackers.} We tune four related tracking-by-detection methods with increasingly
different association and motion mechanisms. SORT \citep{bewley2016sort} provides the
basic pipeline: a Kalman filter predicts track locations, detections are associated to
tracks using Intersection-over-Union and the Hungarian algorithm, and lifecycle parameters
control track retention and confirmation. ByteTrack \citep{zhang2022bytetrack} extends
this pipeline with a second association stage over low-confidence detections. SparseTrack
\citep{liu2023sparsetrack} further partitions detections by pseudo-depth and performs
association stratum by stratum within this two-stage scheme. MoveSORT-KF
\citep{adzemovic2024movesort} modifies both motion modelling and association by
incorporating camera-motion compensation from BoT-SORT \citep{aharon2022botsort} and
replacing the IoU cost with the Move cost. Together, these trackers form a related family
of tracking-by-detection pipelines while exposing different sets of tunable association,
motion, and lifecycle parameters. We restrict the study to this family and therefore do
not include appearance-based or observation-centric variants such as Deep SORT
\citep{wojke2017deepsort} and OC-SORT \citep{cao2023ocsort}.

\textbf{Object detection.} We use the public YOLOX-X \citep{ge2021yolox} detectors
released
with each dataset. The detector is frozen and identical across every configuration,
optimizer, and tracker
within a dataset, so differences between the runs reported here originate in the tracker
configuration. Tracker outputs are postprocessed before scoring, following the pipelines
used in the
tracker papers. Detector configurations and postprocessing settings are given in
Appendix~\ref{app:detector} and Appendix~\ref{app:implementation}.

\textbf{Optimization methods.} Table~\ref{tab:optimizers} lists the eight methods we
consider. They span six families: model-free random search, the two model-based samplers
TPE
and GP-BO, the two multi-fidelity schedules Hyperband and BOHB, and greedy coordinate
search,
of which MFGCS is the multi-fidelity version. Two of the eight are variants rather than
separate families, namely TPE warm-started from a prior and coordinate search without the
multi-fidelity stage. Section~\ref{sec:optimizer_selection} selects two of the eight for
the
full experiment. Every method operates on the same search space for a given tracker, and
on
the same space for both datasets, so the comparison isolates the search strategy. Each run
is
capped at $100$ full-fidelity trials, which is a stopping condition rather than an equal
computational budget, since cost per trial differs across methods. The settings of each
method
and the full search spaces are listed in Appendix~\ref{app:optimizer_configs} and
Appendix~\ref{app:search_spaces}.

\begin{table}[t]
\centering
\small
\setlength{\tabcolsep}{4pt}
\caption{The eight optimization methods considered, spanning six families. Actual cost is
the cumulative scene count of Equation~\ref{eq:budget}, reported with the results. The two
carried into the cross-tracker comparison are in bold.}
\label{tab:optimizers}
\begin{tabular}{lll}
\toprule
Optimizer & Family & Key settings \\
\midrule
Random                & Model-free          & uniform sampling \\
\best{TPE}            & \best{Model-based}  & \best{$\gamma = 0.20$, 24 EI candidates} \\
TPE + prior           & Model-based         & warm-started from the manual config \\
GP-BO                 & Model-based         & expected improvement (EI), 20 startup trials \\
Random + Hyperband    & Multi-fidelity      & rung schedule over scene subsets \\
BOHB                  & Multi-fidelity      & TPE sampling with Hyperband pruning \\
GCS                   & Direct              & greedy coordinate search, no subset stage \\
\best{MFGCS}          & \best{Multi-fidelity, direct} & \best{grid, $g=5$, $r=3$, $m=6$, $T=6$} \\
\bottomrule
\end{tabular}
\end{table}

\subsection{Choosing the optimization methods}
\label{sec:optimizer_selection}

Running every optimization method on every tracker--dataset pair would already require $8
\times 4 \times 2 = 64$ optimization runs, each taking approximately two to seven hours on
the hardware described in Section~\ref{sec:setup}. More importantly, this count assumes a
single fixed configuration for each optimizer. The optimizers themselves expose
hyperparameters that must be selected to ensure a fair comparison; even a modest average
of 16 candidate settings per optimizer would expand the full factorial study to more than
$1000$ optimization runs, corresponding to months of continuous compute. We therefore
adopt a two-stage evaluation protocol. First, we compare all eight optimization methods
under a common search space using SORT on DanceTrack-val. We then select the two strongest
methods from this controlled comparison and evaluate them across all four trackers and
both datasets.

\begin{figure}[t]

\centering
\includegraphics[width=\textwidth]{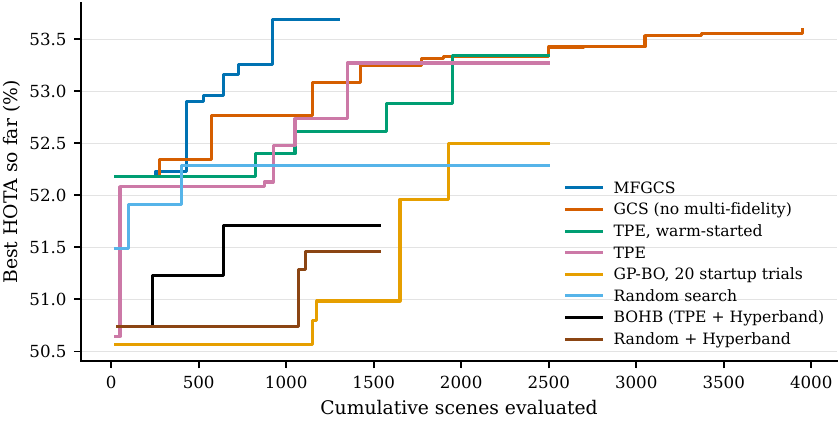}
\caption{Running-best HOTA against cumulative scenes evaluated, on the SORT and
DanceTrack-val search space, for the best configuration of each optimization method.}
\label{fig:families}
\end{figure}

Table~\ref{tab:ablation_families} reports the outcome. The single-fidelity samplers pay
the
full per-trial cost on every candidate and so consume $2500$ scenes for $100$ trials,
whereas
MFGCS reaches a higher HOTA at roughly half that budget. The two Hyperband-pruned methods
are cheaper
than those single-fidelity samplers, at about $1530$ scenes, but score lowest of the
eight; we
return to them in Appendix~\ref{app:negative_results}. MFGCS records the highest observed
score at $53.69\%$. The highest-scoring sampler is TPE warm-started from the manually
tuned
configuration at $53.35\%$, which we tried out of curiosity about whether a prior helps in
general, with plain TPE just behind at $53.27\%$. We select MFGCS and plain TPE for the
broader experiment, because warm-starting would make the sampler depend on the hand-tuned
configuration the benchmark exists to replace. Figure~\ref{fig:families} shows more
detailed results as running-best curves, with MFGCS in
the upper-left region: it reaches a higher score than any other method at every budget
beyond
its first few hundred scenes.

\begin{table}[t]
\centering
\footnotesize
\setlength{\tabcolsep}{4pt}
\caption{Best result per optimization method on the SORT and DanceTrack-val search space.
Scenes is the cumulative count of Equation~\ref{eq:budget} and is the cost axis; trials
are bookkeeping. The highest HOTA is in bold. TPE and MFGCS are the two methods carried
into the cross-tracker experiment, for the reasons given in the text.}
\label{tab:ablation_families}
\begin{tabular}{lrrr}
\toprule
Method & HOTA & Trials & Scenes \\
\midrule
Random + Hyperband & 51.46 & 100 & 1534 \\
BOHB               & 51.71 & 100 & 1530 \\
Random             & 52.28 & 100 & 2500 \\
GP-BO              & 52.50 & 100 & 2500 \\
TPE                & 53.27 & 100 & 2500 \\
TPE + prior        & 53.35 & 100 & 2500 \\
GCS                & 53.59 & 12  & 3950 \\
\best{MFGCS}       & \best{53.69} & \best{17} & \best{1297} \\
\bottomrule
\end{tabular}
\end{table}

\subsection{Main results}
\label{sec:main_results}

Each tracker is paired with the better of TPE and MFGCS on its validation split, and the
selected configuration is then evaluated unchanged on the test split and compared against
the
number published for that same tracker on that same split, with our metrics returned by
the
official evaluation servers. Table~\ref{tab:test_results} reports the outcome: every
published
baseline is beaten, by up to $6.15$ HOTA points on DanceTrack and $8.69$ on SportsMOT. For
SORT, tuning improves both association accuracy, from $32.60$ to $36.89$ AssA, and
detection
accuracy, from $72.00$ to $79.41$ DetA, lifting HOTA from a published $47.90$ to $54.05$
\citep{adzemovic2024movesort}. The search includes detection-selection thresholds as well
as
association parameters, so both terms are within its reach.

\begin{table}[t]
\centering
\small
\setlength{\tabcolsep}{4pt}
\caption{Test-split results for all eight tuned configurations, each tracker paired with
the better of TPE and MFGCS on its validation split. Our metrics are returned by the
official evaluation servers. An em dash marks a tracker with no published test-split
number. Published values are taken from \citet{sun2022dancetrack} for SORT and ByteTrack
on DanceTrack, \citet{liu2023sparsetrack} for SparseTrack, \citet{adzemovic2024movesort}
for MoveSORT-KF, and \citet{cui2023sportsmot} for ByteTrack on SportsMOT.}
\label{tab:test_results}
\begin{tabular}{llrrrrrrrr}
\toprule
& & \multicolumn{6}{c}{Tuned, this work} & \multicolumn{2}{c}{Published} \\
\cmidrule(lr){3-8}\cmidrule(lr){9-10}
Tracker & Method & HOTA & DetA & AssA & MOTA & IDF1 & IDSW & HOTA & DetA \\
\midrule
\multicolumn{10}{l}{\textit{DanceTrack-test}} \\
SORT        & MFGCS & 54.05 & 79.41 & 36.89 & 90.50 & 55.14 & 1728 & 47.90 & 72.00 \\
ByteTrack   & TPE   & 53.73 & 80.71 & 35.88 & 92.13 & 54.70 & 1964 & 47.70 & 71.00 \\
SparseTrack & MFGCS & 58.17 & 81.94 & 41.44 & 91.42 & 56.68 & 1646 & 55.50 & 78.90 \\
MoveSORT-KF & TPE   & 55.58 & 81.30 & 38.13 & 91.78 & 54.89 & 1716 & 53.30 & 80.70 \\
\midrule
\multicolumn{10}{l}{\textit{SportsMOT-test}} \\
SORT        & TPE   & 72.38 & 87.74 & 59.78 & 96.32 & 72.08 & 2724 & \notavail & \notavail \\
ByteTrack   & MFGCS & 72.79 & 87.94 & 60.32 & 96.57 & 72.68 & 2621 & 64.10 & 78.50 \\
SparseTrack & TPE   & 69.52 & 87.37 & 55.39 & 96.02 & 68.46 & 3591 & \notavail & \notavail \\
MoveSORT-KF & MFGCS & 72.89 & 88.08 & 60.38 & 96.54 & 72.57 & 2617 & 71.00 & 86.20 \\
\bottomrule
\end{tabular}
\end{table}

The same improvement is visible on the validation splits. Table~\ref{tab:main_results}
shows
that both optimization methods outperform every published score and every configuration we
tuned by hand. MFGCS reaches the per-tracker target faster than TPE on seven of the eight
combinations.\footnote{The targets are fixed before the runs and shared by both optimizers
within a cell, each a round HOTA value that both of them clear: $53.0$ for SORT and
ByteTrack, $57.0$ for SparseTrack and $54.0$ for MoveSORT-KF on DanceTrack; $78.0$ for
SORT,
ByteTrack and SparseTrack and $79.0$ for MoveSORT-KF on SportsMOT.} The gains are not
specific to one setting: both methods
outperform the manual baseline in all eight validation settings, and on DanceTrack the
validation-selected configuration moves by at most $0.58$ HOTA points when it is evaluated
on
the test split. On SportsMOT all four score lower on the test split than on the validation
split, by $6.06$ to $8.83$ points, with the largest drop on SparseTrack.

\begin{table}[t]
\centering
\small
\setlength{\tabcolsep}{4pt}
\caption{Validation results. \emph{Paper} is the HOTA published in each tracker's original
work and \emph{Manual} a configuration set by hand. The better of the two optimizers is in
bold within each pair of columns. An em dash marks a cell with no published number.
Published values have the same sources as in Table~\ref{tab:test_results}.}
\label{tab:main_results}
\begin{tabular}{lrrrrrr}
\toprule
& \multicolumn{4}{c}{HOTA} & \multicolumn{2}{c}{Time-to-target} \\
\cmidrule(lr){2-5}\cmidrule(lr){6-7}
Tracker & Paper & Manual & TPE & MFGCS & TPE & MFGCS \\
\midrule
\multicolumn{7}{l}{\emph{DanceTrack-val}} \\
SORT        & 47.80 & 52.18 & 53.27 & \best{53.69} & 125 min & \best{73 min} \\
ByteTrack   & 47.10 & 53.05 & \best{53.64} & 53.29 & 78 min & \best{2.5 min} \\
SparseTrack & 53.90 & 54.37 & 57.34 & \best{58.75} & 207 min & \best{78 min} \\
MoveSORT-KF & 53.30 & 53.35 & \best{55.16} & 54.75 & 97 min & \best{19 min} \\
\midrule
\multicolumn{7}{l}{\emph{SportsMOT-val}} \\
SORT        & \notavail & 74.91 & \best{79.14} & 78.36 & 52 min & \best{51 min} \\
ByteTrack   & 62.80 & 75.88 & 78.85 & 78.85 & \best{30 min} & 110 min \\
SparseTrack & \notavail & 75.56 & \best{78.35} & 78.20 & 195 min & \best{118 min} \\
MoveSORT-KF & \notavail & 76.99 & 79.67 & \best{80.19} & 77 min & \best{48 min} \\
\bottomrule
\end{tabular}
\end{table}

\subsection{Ablation studies}
\label{sec:ablations}

We ablate the scene sampler and the coordinate optimizer of
Section~\ref{sec:mfgcs_components} on the SORT and DanceTrack-val search space, at the same
budget and seed as above. Table~\ref{tab:ablation_mfgcs} reports the sweep and
Figure~\ref{fig:mfgcs_ablation_budget} shows the corresponding running-best curves. The
per-coordinate early-stopping rule is held fixed across every run reported here, so we do not
ablate it.

\textbf{Coordinate optimizer.} We compare three ways of proposing candidate values for one
hyperparameter. Grid search evaluates $g$ evenly spaced points across the current interval,
including both endpoints, and contracts around the best of them. Ternary section instead
evaluates two interior points per round and discards the outer third on the losing side,
which halves the evaluations per round but can never return a value on a boundary. The random
variant draws a fixed number of candidate values uniformly from the interval, with no
contraction at all, and serves as the unstructured control. Appendix~\ref{app:ternary_vs_grid}
derives the evaluation-cost and contraction-rate trade-off between grid and ternary.

Grid search at \texttt{g5r3-n6-s6} ($g = 5$ points per round, $r = 3$ contraction rounds,
$m = 6$ subset scenes, $T = 6$ sweeps) reaches $53.69\%$ HOTA, ahead of the ternary variant at
$53.19\%$ and the random variant at $52.90\%$. We carry grid forward because it scored highest
among the variants we tested here, and because its candidate set includes the interval
endpoints, where several of these hyperparameters take their best observed values. Within grid
search, the coarsest variant we tested did not improve on its starting configuration, because
its lattice re-quantized the best point onto the value the search began from, and three
contraction rounds beat two by $1.1$ HOTA points, whereas resolution past $g = 5$ does not
help.

\textbf{Scene sampler.} The subset size trades comparison noise against budget, and the best
observed setting depends on the coordinate optimizer. Among the sizes we tested, grid does
best at $m = 6$, which reaches $53.69\%$: dropping to $m = 3$ costs $0.74$ points ($52.95\%$)
because the noisier comparison discards good candidates before they reach a full evaluation,
while raising it to $m = 9$ gives back $53.33\%$ for $1280$ scenes, buying less variance
reduction than it costs in budget. For ternary the best observed setting is $m = 3$ at
$53.19\%$, against $52.40\%$ at $m = 6$.

Setting $m$ to all $25$ scenes removes the multi-fidelity stage altogether, since every
candidate is then scored on the full split. That run reaches $53.59\%$ HOTA and consumes
$3950$ cumulative scenes, against MFGCS's $53.69\%$ at $1297$; the $3950$ scenes are $158$
full-split equivalents, despite the run recording only $12$ trials. Comparing the two end
points understates the subset stage, because it allows the full-fidelity variant three times
the budget. Read at matched cost, the full-fidelity variant stands at $53.08\%$ after $1297$
scenes, so the subset stage is worth $0.61$ HOTA points at equal budget as well as cutting
that budget threefold.

\begin{table}[t]
\centering
\footnotesize
\setlength{\tabcolsep}{4pt}
\caption{MFGCS configuration ablation on the SORT and DanceTrack-val search space. The
suffix \texttt{g\{N\}r\{R\}-n\{S\}-s\{T\}} denotes grid points per coordinate, contraction
rounds, subset scenes, and maximum sweeps; \texttt{tern\{N\}} is the ternary variant and
\texttt{rand\{M\}} the random variant with $M$ candidates per coordinate. Grouped by
coordinate optimizer, by descending HOTA within each group; the selected configuration is
in bold.}
\label{tab:ablation_mfgcs}
\begin{tabular}{lrrp{5.6cm}}
\toprule
Configuration & HOTA & Scenes & Observation \\
\midrule
\multicolumn{4}{l}{\emph{Grid}} \\
\best{\texttt{g5r3-n6-s6}} & \best{53.69} & \best{1297} & \best{selected} \\
\texttt{g5r3-n9-s6}   & 53.33 & 1280 & subset noise already below the win margin \\
\texttt{g7r2-n6-s6}   & 53.32 & 1321 & wider lattice does not beat deeper rounds \\
\texttt{g5r3-n3-s6}   & 52.95 & 874  & subset too small, noisy comparisons \\
\texttt{g7r3-n6-s6}   & 52.74 & 1293 & resolution past $g=5$ over-shrinks \\
\texttt{g5r2-n6-s6}   & 52.55 & 911  & two rounds are not enough \\
\texttt{g4r2-n3-s6}   & 52.18 & 254  & grid too coarse to escape the start \\
\midrule
\multicolumn{4}{l}{\emph{Ternary}} \\
\texttt{tern5-n3-s6}  & 53.19 & 997  & best ternary setting \\
\texttt{tern5-n6-s6}  & 52.40 & 1198 & larger subset hurts the ternary variant \\
\midrule
\multicolumn{4}{l}{\emph{Random}} \\
\texttt{rand20-n6-s6} & 52.90 & 1553 & random improves at high candidate count \\
\texttt{rand5-n3-s6}  & 52.18 & 221  & too few random candidates \\
\bottomrule
\end{tabular}
\end{table}

\section{Conclusion}
\label{sec:conclusion}

We benchmarked hyperparameter optimization for tracking-by-detection multi-object
tracking,
comparing eight optimization methods across six families. Within each tracker and dataset
pair they share a search space and an evaluation pipeline, so the comparison isolates the
search strategy. We introduced Multi-Fidelity Greedy Coordinate Search, an optimizer
matched to
the cost structure of the problem. Our main observations are as follows:

\begin{itemize}
  \item The configurations these trackers are commonly run with are under-tuned on both
        benchmarks. On the validation splits both optimization methods outperform every
published score and every configuration we tuned by hand, on all four trackers. Taking
both
optimizers across all cells, the margin over the published scores runs from $1.45$ to
$16.05$ HOTA points, and over our own hand-tuned configurations from $0.24$ to $4.38$
        points. For each setting the better
        validation-selected configuration also outperforms the published number on the
        held-out test split, by $1.9$ to $8.7$ points.
  \item Hand-tuning consumed a few hours to a full day per tracker and scored below both
optimizers on every row, while MFGCS reached a higher-scoring configuration in under
        two hours of unattended wall-clock.
  \item MFGCS and TPE reach comparable HOTA in these runs, differing by at most $1.41$
        points in
any cell and by under $0.8$ in seven of the eight, and MFGCS reaches the per-tracker
target faster on seven of the eight, and its scene-subset stage alone cuts the
budget threefold while scoring $0.61$ HOTA points above the full-fidelity coordinate
        search at equal cost. The reduction has to be paired with the search to pay off.
        Hyperband and BOHB reduce fidelity across independently sampled configurations
        instead, each scored on its own scene sample, and pruning on that unpaired signal
        cost $0.82$ and $1.56$ HOTA points against the same samplers without it.
\end{itemize}

Our results show that a tracker's reported number and its achievable number are different
quantities, and that the difference is large enough to change how two published methods
compare. We release the optimized configurations so that a tuned baseline is available to
compare against.

\appendix
\section{Experimental details}
\label{app:experimental_details}

This section provides the configurations summarized in Section~\ref{sec:setup}. We first
describe the object detectors, then the inference and postprocessing pipeline, then the
search spaces, and finally the settings of each optimizer.

\subsection{Object detection}
\label{app:detector}

Both datasets use a YOLOX-X detector \citep{ge2021yolox}. For DanceTrack we use the
YOLOX-X checkpoint released with the ByteTrack ablation setup, and for SportsMOT the
YOLOX-X checkpoint released with the dataset. Because the detector is frozen, detections
are computed once per dataset and written to a cache addressed by scene name and frame
index, so every trial reads identical detections and the objective stays deterministic.
A warm-cache pass over DanceTrack-val costs two to three minutes, against roughly $1.5$
hours with the detector in the loop.

\subsection{Tracker inference postprocessing}
\label{app:implementation}

Tracker outputs are postprocessed before scoring using trajectory interpolation methods,
with settings that differ per dataset. Gaps are filled by linear interpolation when the
track is lost for at most $10$ frames on DanceTrack, or $20$ on SportsMOT, and the
surrounding tracklet is at least $30$ frames long. Final tracks shorter than $20$ frames
on DanceTrack, or $10$ on SportsMOT, are removed. Every score reported in this paper,
including every optimizer trial, is computed on the postprocessed output.

\subsection{Search spaces}
\label{app:search_spaces}

Table~\ref{tab:search_space} lists every hyperparameter that is part of the tracker's
search space. The shared block applies to all four trackers; the remaining blocks are
added per tracker. This gives $7$ hyperparameters for SORT, $9$ for MoveSORT-KF, $12$ for
ByteTrack, and $14$ for SparseTrack, so the spaces differ in size by a factor of two
across the four methods.

\begin{table}[ht]
\centering
\footnotesize
\setlength{\tabcolsep}{5pt}
\caption{Search spaces. Ranges are inclusive. $\dagger$ marks a hyperparameter whose
candidates are generated at or above the sampled detection threshold. This constrains how
candidates are proposed and is not an invariant of every stored configuration: a
coordinate search that accepts no move for this parameter retains its existing value.
$\ddagger$ marks a logarithmic scale.}
\label{tab:search_space}
\begin{tabular}{llll}
\toprule
Tracker & Hyperparameter & Type & Range \\
\midrule
\multirow{4}{*}{All}
 & detection threshold                  & float & $[0.1, 0.9]$ \\
 & initialization threshold             & int   & $[0, 3]$ \\
 & remember threshold                   & int   & $[1, 60]$ \\
 & motion-model position noise scale    & float & $[0.01, 0.25]$ \\
\midrule
\multirow{3}{*}{SORT}
 & new-track detection threshold$^{\dagger}$ & float       & $[0.3, 0.9]$ \\
 & IoU match threshold                       & float       & $[0.10, 0.50]$ \\
 & score fusion                              & categorical & $\{\text{on}, \text{off}\}$ \\
\midrule
\multirow{8}{*}{ByteTrack}
 & new-track detection threshold$^{\dagger}$ & float       & $[0.1, 0.9]$ \\
 & duplicate IoU threshold                   & float       & $[0.5, 1.0]$ \\
 & high-tier match threshold                 & float       & $[0.10, 0.50]$ \\
 & high-tier score fusion                    & categorical & $\{\text{on}, \text{off}\}$ \\
 & low-tier match threshold                  & float       & $[0.30, 0.70]$ \\
 & low-tier score fusion                     & categorical & $\{\text{on}, \text{off}\}$ \\
 & new-track match threshold                 & float       & $[0.10, 0.50]$ \\
 & new-track score fusion                    & categorical & $\{\text{on}, \text{off}\}$ \\
\midrule
\multirow{10}{*}{SparseTrack}
 & new-track detection threshold$^{\dagger}$ & float       & $[0.1, 0.9]$ \\
 & duplicate IoU threshold                   & float       & $[0.5, 1.0]$ \\
 & high-tier match threshold                 & float       & $[0.10, 0.50]$ \\
 & high-tier score fusion                    & categorical & $\{\text{on}, \text{off}\}$ \\
 & high-tier depth levels                    & int         & $[6, 20]$ \\
 & low-tier match threshold                  & float       & $[0.30, 0.70]$ \\
 & low-tier score fusion                     & categorical & $\{\text{on}, \text{off}\}$ \\
 & low-tier depth levels                     & int         & $[6, 20]$ \\
 & new-track match threshold                 & float       & $[0.10, 0.50]$ \\
 & new-track score fusion                    & categorical & $\{\text{on}, \text{off}\}$ \\
\midrule
\multirow{5}{*}{MoveSORT-KF}
 & new-track detection threshold$^{\dagger}$ & float       & $[0.1, 0.9]$ \\
 & Move match threshold                      & float       & $[0.10, 0.50]$ \\
 & score fusion                              & categorical & $\{\text{on}, \text{off}\}$ \\
 & motion weight $\lambda^{\ddagger}$        & float       & $[1.0, 15.0]$ \\
 & motion distance                           & categorical & $\{\ell_1, \ell_2\}$ \\
\bottomrule
\end{tabular}
\end{table}

\subsection{Optimizer configurations}
\label{app:optimizer_configs}

Every optimizer except MFGCS is used from the Optuna framework \citep{akiba2019optuna}.
The TPE sampler is Optuna's \texttt{TPESampler} with a univariate kernel, $24$
expected-improvement candidates per step, and a good-or-bad split quantile of
$\gamma = 0.20$. The quantile is a setting we selected, not an Optuna default; Optuna's
own
default is $\min(\lceil 0.1n \rceil, 25)$. Optuna's \texttt{GPSampler} uses expected
improvement with $20$ startup trials in the configuration reported in
Table~\ref{tab:ablation_families}, and $10$ in the default variant, and the Hyperband
variants use a rung schedule over scene subsets
with a reduction factor of $3$. Random search samples uniformly over the same space. Every
optimizer uses seed $42$ and a cap of $100$ full-fidelity evaluations.
Table~\ref{tab:mfgcs_defaults} gives the MFGCS settings.

\begin{table}[ht]
\centering
\small
\setlength{\tabcolsep}{4pt}
\caption{MFGCS default configuration, abbreviated \texttt{g5r3-n6-s6} in the main text.}
\label{tab:mfgcs_defaults}
\begin{tabular}{llp{5.9cm}}
\toprule
Parameter & Default & Purpose \\
\midrule
coordinate optimizer & grid & Coarse-to-fine grid search per coordinate. \\
$g$, grid points     & 5    & Points per round, including both endpoints. \\
$r$, rounds          & 3    & Contraction rounds per coordinate. \\
scene sampler        & uniform random & Draws the low-fidelity subset. \\
$m$, subset size     & 6    & Scenes drawn per coordinate at low fidelity. \\
$T$, max sweeps      & 6    & Outer-loop cap. \\
max trials           & 100  & Hard cap on full-fidelity evaluations. \\
early stop           & on   & Stop after a sweep with no accepted move. \\
$\varepsilon$        & $10^{-4}$ & Minimum HOTA gain to accept, at the noise floor $\sigma_d / \sqrt{m}$. \\
barren-sweep drop    & 2    & Sweeps without an accept before a coordinate is dropped. \\
$\rho$, shrink       & 0.25 & Half-width of the post-accept interval, as a fraction of the current width. \\
bootstrap            & on   & One full evaluation of the initial configuration seeds $v$. \\
\bottomrule
\end{tabular}
\end{table}

\section{Additional results}
\label{app:additional_results}

This section reports the full metric panel behind Table~\ref{tab:main_results}, the
optimized hyperparameter values themselves, and the information needed to reproduce them.

\subsection{Full metrics and optimized hyperparameters}
\label{app:full_metrics}

Table~\ref{tab:full_metrics} reports every metric at the best-HOTA trial each optimizer
selected. The winner on HOTA is not always the winner on identity switches: on DanceTrack
MoveSORT-KF, TPE leads on HOTA, DetA, AssA, MOTA and IDF1 while MFGCS commits $98$ fewer
identity switches.

\begin{table}[ht]
\centering
\footnotesize
\setlength{\tabcolsep}{4pt}
\caption{Validation metrics at the best-HOTA trial; better of the two optimizers in bold.}
\label{tab:full_metrics}
\begin{tabular}{llrrrrrrr}
\toprule
Tracker & Method & HOTA & DetA & AssA & MOTA & IDF1 & IDSW & FPS \\
\midrule
\multicolumn{9}{l}{\emph{DanceTrack-val}} \\
\multirow{2}{*}{SORT}
 & TPE   & 53.27 & \best{78.33} & 36.36 & \best{89.86} & 53.26 & 1669 & \multirow{2}{*}{930--1082} \\
 & MFGCS & \best{53.69} & 76.47 & \best{37.81} & 87.87 & \best{55.60} & \best{1573} & \\
\multirow{2}{*}{ByteTrack}
 & TPE   & \best{53.64} & 78.60 & \best{36.77} & 90.52 & \best{54.14} & 1842 & \multirow{2}{*}{1012--1121} \\
 & MFGCS & 53.29 & \best{79.08} & 36.04 & \best{90.66} & 53.68 & \best{1652} & \\
\multirow{2}{*}{SparseTrack}
 & TPE   & 57.34 & 77.32 & 42.65 & 87.12 & 58.52 & 1584 & \multirow{2}{*}{698--755} \\
 & MFGCS & \best{58.75} & \best{78.86} & \best{43.95} & \best{89.67} & \best{58.73} & \best{1524} & \\
\multirow{2}{*}{MoveSORT-KF}
 & TPE   & \best{55.16} & \best{78.61} & \best{38.87} & \best{90.27} & \best{54.96} & 1559 & \multirow{2}{*}{845--964} \\
 & MFGCS & 54.75 & 78.06 & 38.53 & 88.72 & 54.50 & \best{1461} & \\
\midrule
\multicolumn{9}{l}{\emph{SportsMOT-val}} \\
\multirow{2}{*}{SORT}
 & TPE   & \best{79.14} & 92.14 & \best{67.99} & 98.73 & \best{79.14} & \best{625} & \multirow{2}{*}{930--1082} \\
 & MFGCS & 78.36 & \best{92.38} & 66.49 & \best{98.80} & 77.93 & 660 & \\
\multirow{2}{*}{ByteTrack}
 & TPE   & 78.85 & \best{92.36} & 67.34 & 98.81 & 78.43 & 629 & \multirow{2}{*}{1012--1121} \\
 & MFGCS & 78.85 & 92.29 & \best{67.39} & \best{98.83} & \best{78.78} & \best{625} & \\
\multirow{2}{*}{SparseTrack}
 & TPE   & \best{78.35} & \best{92.11} & \best{66.66} & 98.45 & \best{78.32} & \best{731} & \multirow{2}{*}{698--755} \\
 & MFGCS & 78.20 & 91.98 & 66.51 & \best{98.53} & 78.03 & 827 & \\
\multirow{2}{*}{MoveSORT-KF}
 & TPE   & 79.67 & 92.31 & 68.78 & 98.64 & 79.56 & 595 & \multirow{2}{*}{845--964} \\
 & MFGCS & \best{80.19} & \best{92.48} & \best{69.55} & \best{98.83} & \best{80.27} & \best{552} & \\
\bottomrule
\end{tabular}
\end{table}

\label{app:best_params}

Table~\ref{tab:best_sort_movesort} and Table~\ref{tab:best_byte_sparse} give the stored
configuration values each optimizer selected, i.e. the values that produce the HOTA scores
of
Table~\ref{tab:main_results}. For SORT and MoveSORT-KF the tracker first filters
detections
by the detection threshold and only then applies the new-track gate, so the effective
confidence required to start a track is the larger of the two. The DanceTrack MoveSORT-KF
column stores $0.75$ and $0.70$, for instance, giving an effective threshold of $0.75$;
the
stored $0.70$ gate admits nothing further. The searches selected substantially different
values between datasets for the same tracker.

\begin{table}[ht]
\centering
\small
\setlength{\tabcolsep}{4pt}
\caption{Optimized SORT and MoveSORT-KF hyperparameters. The MoveSORT-KF SportsMOT MFGCS
column is the configuration evaluated on SportsMOT-test in Table~\ref{tab:test_results}.}
\label{tab:best_sort_movesort}
\begin{tabular}{lrrrr}
\toprule
 & \multicolumn{2}{c}{DanceTrack} & \multicolumn{2}{c}{SportsMOT} \\
\cmidrule(lr){2-3}\cmidrule(lr){4-5}
Hyperparameter & TPE & MFGCS & TPE & MFGCS \\
\midrule
\multicolumn{5}{l}{\emph{SORT}} \\
detection threshold & 0.5498 & 0.2 & 0.186 & 0.6 \\
initialization threshold & 3 & 3 & 1 & 3 \\
remember threshold & 25 & 28 & 43 & 30 \\
position noise scale & 0.0257 & 0.05 & 0.0144 & 0.025 \\
match threshold & 0.2254 & 0.325 & 0.1288 & 0.1 \\
score fusion & off & off & on & off \\
new-track detection threshold & 0.8417 & 0.9 & 0.8147 & 0.6562 \\
HOTA & 53.27 & 53.69 & 79.14 & 78.36 \\
\midrule
\multicolumn{5}{l}{\emph{MoveSORT-KF}} \\
detection threshold & 0.648 & 0.75 & 0.7337 & 0.675 \\
initialization threshold & 1 & 0 & 3 & 3 \\
remember threshold & 25 & 30 & 49 & 30 \\
position noise scale & 0.0373 & 0.05 & 0.0171 & 0.01 \\
match threshold & 0.204 & 0.2 & 0.145 & 0.1 \\
score fusion & off & off & off & off \\
motion weight $\lambda$ & 7.7777 & 3.873 & 5.6847 & 7.622 \\
motion distance & $\ell_2$ & $\ell_1$ & $\ell_1$ & $\ell_1$ \\
new-track detection threshold & 0.7454 & 0.7 & 0.7693 & 0.7 \\
HOTA & 55.16 & 54.75 & 79.67 & 80.19 \\
\bottomrule
\end{tabular}
\end{table}

\begin{table}[ht]
\centering
\small
\setlength{\tabcolsep}{4pt}
\caption{Optimized ByteTrack and SparseTrack hyperparameters. The SparseTrack DanceTrack
MFGCS column is the configuration evaluated on DanceTrack-test in
Table~\ref{tab:test_results}.}
\label{tab:best_byte_sparse}
\begin{tabular}{lrrrr}
\toprule
 & \multicolumn{2}{c}{DanceTrack} & \multicolumn{2}{c}{SportsMOT} \\
\cmidrule(lr){2-3}\cmidrule(lr){4-5}
Hyperparameter & TPE & MFGCS & TPE & MFGCS \\
\midrule
\multicolumn{5}{l}{\emph{ByteTrack}} \\
detection threshold & 0.5254 & 0.6 & 0.3443 & 0.55 \\
initialization threshold & 3 & 3 & 0 & 3 \\
remember threshold & 21 & 19 & 6 & 30 \\
position noise scale & 0.0793 & 0.05 & 0.0158 & 0.0225 \\
duplicate IoU threshold & 0.9996 & 1 & 0.977 & 1 \\
high-tier match threshold & 0.1987 & 0.25 & 0.1001 & 0.1 \\
high-tier score fusion & off & off & on & off \\
low-tier match threshold & 0.313 & 0.5 & 0.6266 & 0.45 \\
low-tier score fusion & on & off & on & off \\
new-track match threshold & 0.4264 & 0.25 & 0.1043 & 0.1 \\
new-track score fusion & on & off & off & off \\
new-track detection threshold & 0.6972 & 0.7 & 0.3818 & 0.6812 \\
HOTA & 53.64 & 53.29 & 78.85 & 78.85 \\
\midrule
\multicolumn{5}{l}{\emph{SparseTrack}} \\
detection threshold & 0.6439 & 0.6 & 0.7308 & 0.4 \\
initialization threshold & 0 & 3 & 0 & 3 \\
remember threshold & 24 & 14 & 14 & 10 \\
position noise scale & 0.2387 & 0.2172 & 0.01 & 0.05 \\
duplicate IoU threshold & 0.8567 & 0.9062 & 0.9987 & 1 \\
high-tier match threshold & 0.2804 & 0.35 & 0.1235 & 0.1063 \\
high-tier score fusion & off & off & off & off \\
high-tier depth levels & 16 & 12 & 17 & 12 \\
low-tier match threshold & 0.6909 & 0.65 & 0.473 & 0.7 \\
low-tier score fusion & off & off & off & off \\
low-tier depth levels & 20 & 16 & 7 & 12 \\
new-track match threshold & 0.4122 & 0.25 & 0.2371 & 0.1063 \\
new-track score fusion & on & off & on & off \\
new-track detection threshold & 0.8937 & 0.8062 & 0.7393 & 0.7 \\
HOTA & 57.34 & 58.75 & 78.35 & 78.20 \\
\bottomrule
\end{tabular}
\end{table}

\section{Additional ablations}
\label{app:additional_ablations}

This section expands the ablation findings of Section~\ref{sec:ablations}: the MFGCS
configuration sweep, the configurations of the baseline optimizers, and the formal
comparison
of the two contractive coordinate optimizers.

\subsection{MFGCS configuration}
\label{app:mfgcs_ablation}

\begin{figure}[ht]
\centering
\includegraphics[width=\textwidth]{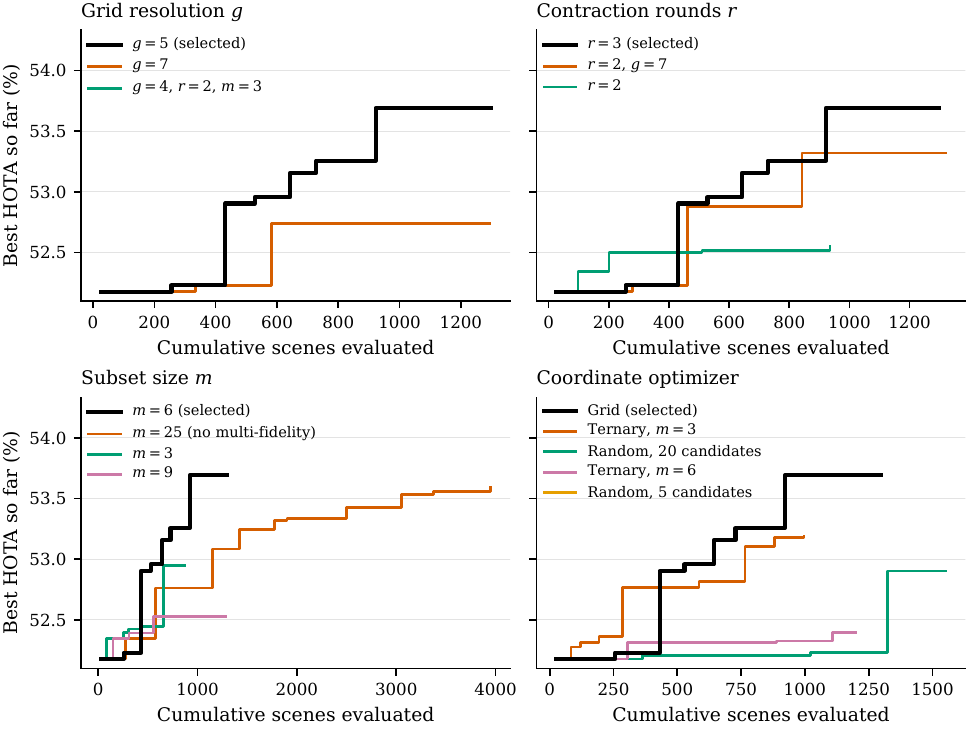}
\caption{MFGCS configuration ablation on the SORT and DanceTrack-val search space. Each
panel varies one setting and repeats the selected configuration in black as the common
reference. Note the differing horizontal scales: the $m = 25$ run evaluates every scene at
every step and so spends roughly three times the budget of the others.}
\label{fig:mfgcs_ablation_budget}
\end{figure}

Table~\ref{tab:ablation_mfgcs} reports the sweep over the MFGCS coordinate-search and
subset-sampling parameters, and Figure~\ref{fig:mfgcs_ablation_budget} shows the corresponding running-best curves. Grid
search gives the best configuration overall, at $53.69\%$
against $53.19\%$ for the best ternary variant and $52.90\%$ for the best random one,
which needs $20$ candidates per coordinate and $1553$ scenes to get there. The setting
that matters most is the grid resolution $g$, because too coarse a lattice re-quantizes
the optimum onto the point the search began from and the sweep never moves:
\texttt{g4r2-n3-s6} ties the untuned baseline at $52.18\%$ after only $254$ scenes. Given
$g = 5$, three contraction rounds beat two by over a point ($53.69\%$ against $52.55\%$)
at a comparable budget, while $g = 7$ does not help at either depth, and the subset size
trades noise against budget in both directions: $m = 3$ discards good candidates before
they reach a full evaluation ($52.95\%$), whereas $m = 9$ spends $1280$ scenes to land
within noise of $m = 6$.

\subsection{Ternary section and grid search comparison}
\label{app:ternary_vs_grid}

The ternary and grid coordinate optimizers of Section~\ref{sec:mfgcs_components} are both
one-dimensional contractive search methods, but they are not equivalent. Ternary section
is not a special case of grid search for any choice of $g$ and $r$.

\subsubsection{Evaluation points}

For a window $[A, B]$ of width $w = B - A$, ternary section
evaluates the two interior points $m_1 = A + w/3$ and $m_2 = B - w/3$, never the
endpoints. Grid search with $g$ points evaluates $g$ evenly spaced points including both
endpoints, so $g = 4$ evaluates $\{A, A + w/3, A + 2w/3, B\}$, i.e. both ternary points
\emph{plus} the endpoints. No setting of $g$ reproduces the ternary evaluation set,
because
grid always includes the endpoints and ternary never does.

\subsubsection{Contraction rule}

Ternary discards the third of the window on the far side of
the better interior point, so the new width is always $\tfrac{2}{3}w$ regardless of where
the optimum lies. Grid recenters on the best point,
$[\,\hat{x} - \delta,\, \hat{x} + \delta\,]$ with $\delta = w/(g-1)$, giving $2w/(g-1)$
for
an interior optimum and $w/(g-1)$ when the optimum is at an endpoint and the window is
clipped. At $g = 4$ that is $\tfrac{2}{3}w$ interior, the same rate as ternary, and
$\tfrac{1}{3}w$ at an endpoint, which is faster.

\subsubsection{Evaluation cost}

Halving the window costs ternary about $3.4$ evaluations and grid
at $g = 4$ about $6.8$ for an interior optimum, but only $2.5$ at an endpoint. Ternary is
therefore roughly twice as cheap for the same contraction rate, \emph{provided} the
optimum
is interior and the objective is strictly unimodal; grid reaches parity or wins otherwise.
On integer coordinates the
two ternary points frequently round to the same lattice point over a short range, which
makes the quantized grid lattice more reliable, and on unordered categoricals both
degenerate to enumeration.

\subsection{Baseline optimizer configurations}
\label{app:tpe_ablation}
\label{app:gp_ablation}
\label{app:negative_results}

\begin{figure}[ht]
\centering
\includegraphics[width=\textwidth]{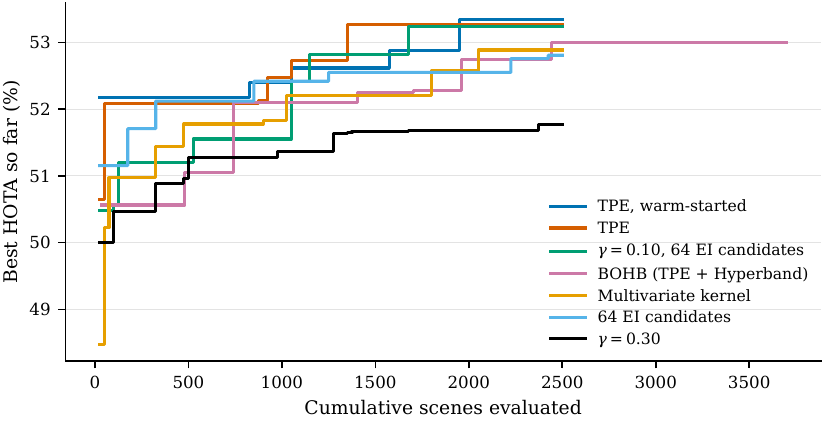}
\caption{Running-best HOTA per TPE variant against cumulative scenes evaluated.}
\label{fig:tpe_ablation_budget}
\end{figure}

\textbf{TPE.} We ran several experiments to tune TPE beyond the configuration we selected,
and none of them improved on it by more than noise, which is why that configuration is the
one carried into the cross-tracker experiment. The warm-started variant records the
highest
score of the TPE family and is the one plotted in Figure~\ref{fig:families}; plain TPE is
the
one selected, because warm-starting depends on the manually tuned configuration. Widening
the candidate pool, switching to the multivariate kernel, and
warm-starting from the manual configuration all land within a point of it, while the only
setting that hurts materially is a split quantile of $\gamma = 0.30$, which loses $1.5$
points because a larger quantile blurs the density ratio driving the acquisition. All
variants in Table~\ref{tab:tpe_ablation} run $100$ trials and consume $2500$ cumulative
scenes, and their reported durations are all approximately $258$ minutes, so wall-clock is
omitted. Figure~\ref{fig:tpe_ablation_budget} shows the running-best curves.

\begin{table}[ht]
\centering
\small
\setlength{\tabcolsep}{4pt}
\caption{TPE configuration ablation on the SORT and DanceTrack-val search space, sorted by
HOTA. Every row is the TPE sampler; the \emph{Sampler} column lists what differs from its
defaults. The variant carried into Section~\ref{sec:main_results} is in bold.}
\label{tab:tpe_ablation}
\begin{tabular}{lp{6.0cm}r}
\toprule
Configuration & Sampler & HOTA \\
\midrule
\texttt{tpe\_sort\_gamma030}        & $\gamma = 0.30$ & 51.77 \\
\texttt{tpe\_sort\_neic64}          & $24 \rightarrow 64$ EI candidates & 52.81 \\
\texttt{tpe\_sort\_multivariate}    & multivariate kernel & 52.88 \\
\texttt{tpe\_sort\_neic64\_gamma010} & $\gamma = 0.10$, $64$ EI candidates & 53.24 \\
\best{\texttt{tpe\_sort}}           & \best{selected ($\gamma = 0.20$, $24$ candidates)} & \best{53.27} \\
\texttt{tpe-prior\_sort}            & warm-started with the manually-tuned config & 53.35 \\
\bottomrule
\end{tabular}
\end{table}

\begin{figure}[ht]
\centering
\includegraphics[width=\textwidth]{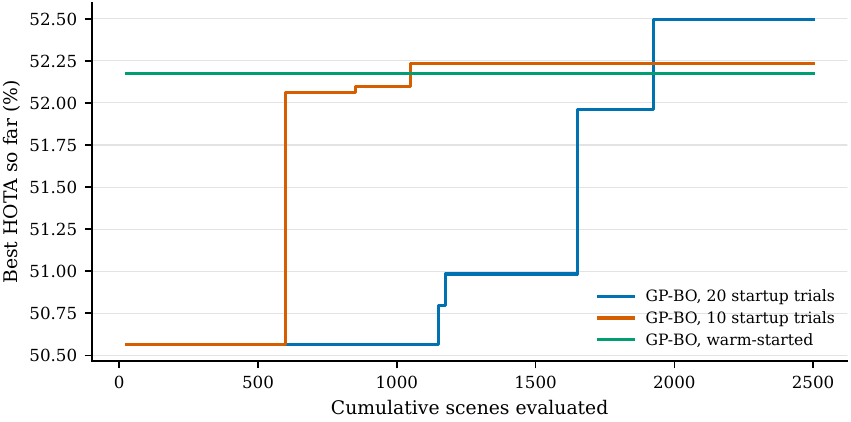}
\caption{Running-best HOTA per Gaussian-process variant against cumulative scenes
evaluated.}
\label{fig:gp_ablation_budget}
\end{figure}

\textbf{Gaussian process.} We likewise varied the GP sampler, and no setting lifted it
clearly above the model-free baseline: two of the three variants in
Table~\ref{tab:gp_ablation} sit below the $52.28\%$ of uniform random search and the third
reaches $52.50\%$, a margin this comparison does not resolve. Doubling the startup trials
scores slightly higher, and seeding the sampler with the manual
configuration scores slightly lower. The three variants run $100$ trials, consume $2500$
cumulative scenes, and take about $259$
minutes; Figure~\ref{fig:gp_ablation_budget} shows the running-best curves.

\begin{table}[ht]
\centering
\small
\setlength{\tabcolsep}{4pt}
\caption{Gaussian-process configuration ablation on the SORT and DanceTrack-val search
space, sorted by HOTA.}
\label{tab:gp_ablation}
\begin{tabular}{lp{6.0cm}r}
\toprule
Configuration & Sampler & HOTA \\
\midrule
\texttt{sort-gp-prior}    & prior-seeded, manually-tuned configuration at trial $0$ & 52.18 \\
\texttt{sort-gp}          & $10$ startup trials & 52.23 \\
\texttt{sort-gp-warmup20} & $20$ startup trials & 52.50 \\
\bottomrule
\end{tabular}
\end{table}

\textbf{Hyperband pruning.} Table~\ref{tab:negative_results} pairs each Hyperband-pruned
method with the same sampler run without pruning, which isolates pruning as a factor
across
two samplers. Pruning works as designed on the budget axis: it removes $78\%$ and $76\%$
of trials and cuts
the cost from $2500$ scenes to $1534$ and $1530$. It also costs accuracy. Read at matched
budget rather than at each run's end point, the unpruned samplers have already reached
$52.28\%$ by $1534$ scenes and $53.27\%$ by $1530$, so pruning costs $0.82$ and $1.56$
HOTA
points at equal cost.

We attribute this to the structure of the low-fidelity comparison rather than to pruning
as
such. Hyperband ranks whole configurations against one another at a rung, and each trial
draws its own scene subset, so the sampled scenes and the configuration change are
confounded and the noise does not cancel. MFGCS reduces fidelity on the same axis but
compares candidate values of a single coordinate, with the others held fixed and all
candidates scored on one shared sample, so the per-scene component is common to both sides
of every comparison. This is an explanation of the observed difference, not a measurement
of
the mechanism, and the pairing is not intrinsic: fixing one subset per rung across trials
would pair Hyperband's comparisons too. Each cell is a single run at seed $42$.

\begin{table}[ht]
\centering
\footnotesize
\setlength{\tabcolsep}{3pt}
\renewcommand{\arraystretch}{1.15}
\caption{Effect of Hyperband pruning, each pruned method paired with the same sampler run
without it, plus the coordinate-greedy run without the multi-fidelity stage. Study names
omit their shared \texttt{sort-} prefix.}
\label{tab:negative_results}
\begin{tabular}{@{}l>{\raggedright\arraybackslash}p{5.2cm}rrr@{}}
\toprule
Study & Algorithm & HOTA & Trials & Scenes \\
\midrule
\texttt{random}           & Random, no pruning & 52.28 & 100 & 2500 \\
\texttt{hyperband-random} & Random + Hyperband & 51.46 & 100 & 1534 \\
\midrule
\texttt{tpe}              & TPE, no pruning    & 53.27 & 100 & 2500 \\
\texttt{bohb}             & TPE + Hyperband    & 51.71 & 100 & 1530 \\
\midrule
\texttt{gcs-no-mf-g5r3-n25-s6} & Coordinate greedy, no multi-fidelity & 53.59 & 12 & 3950 \\
\bottomrule
\end{tabular}
\end{table}

\clearpage

\bibliographystyle{plainnat}
\bibliography{refs}

\end{document}